\documentclass[review]{elsarticle}

\usepackage{lineno,hyperref}
\usepackage{amsmath}
\usepackage{amssymb}
\usepackage{array}
\usepackage[toc,page]{appendix}
\modulolinenumbers[5]
\newcolumntype{C}[1]{>{\centering\arraybackslash}p{#1}}

\journal{Journal of Pattern Recognition}

\begin{document}

\begin{frontmatter}

\title{Relational Long Short-Term Memory\\for Video Action Recognition}

\author[]{Zexi Chen\corref{cor1}}
\ead{zchen22@ncsu.edu}
\cortext[cor1]{Corresponding author}

\author{Bharathkumar Ramachandra}
\ead{bramach2@ncsu.edu}

\author{Tianfu Wu}
\ead{tianfu\_wu@ncsu.edu}

\author{Ranga Raju Vatsavai}
\ead{rrvatsav@ncsu.edu}

\address{North Carolina State University, NC, 27606, USA}
%% Group authors per affiliation:
%\author{Elsevier\fnref{myfootnote}}
%\address{Radarweg 29, Amsterdam}

%% or include affiliations in footnotes:
%\author[mymainaddress,mysecondaryaddress]{Elsevier Inc}
%\ead[url]{www.elsevier.com}

%\author[mysecondaryaddress]{Global Customer Service\corref{mycorrespondingauthor}}
%\cortext[mycorrespondingauthor]{Corresponding author}
%\ead{support@elsevier.com}

%\address[mymainaddress]{1600 John F Kennedy Boulevard, Philadelphia}
%\address[mysecondaryaddress]{360 Park Avenue South, New York}

\begin{abstract}
Spatial and temporal relationships, both short-range and long-range, between objects in videos are key cues for recognizing actions. It is a challenging problem to model them jointly. In this paper, we first present a new variant of Long Short-Term Memory, namely Relational LSTM, to address the challenge of relation reasoning across space and time between objects. In our Relational LSTM module, we utilize a non-local operation similar in spirit to the recently proposed non-local network~\cite{wang2018non} to substitute the fully connected operation in the vanilla LSTM. By doing this, our Relational LSTM is capable of capturing long and short-range spatio-temporal relations between objects in videos in a principled way. Then, we propose a two-branch neural architecture consisting of the Relational LSTM module as the non-local branch and a spatio-temporal pooling based local branch. The local branch is utilized for capturing local spatial appearance and/or short-term motion features. The two branches are concatenated to learn video-level features from snippet-level ones which are then used for classification. Experimental results on UCF-101 and HMDB-51 datasets show that our model achieves state-of-the-art results among LSTM-based methods, while obtaining comparable performance with other state-of-the-art methods (which use not directly comparable schema). Further, on the more complex large-scale Charades dataset we obtain a large 3.2\% gain over state-of-the-art methods, verifying the effectiveness of our method in complex understanding.

\end{abstract}

\begin{keyword}
Relational LSTM, Non-local operator, Spatial-temporal relation reasoning, Action recognition, Video classification
\end{keyword}

\end{frontmatter}

%\linenumbers

\section{Introduction}
Action recognition (or video classification) as we use it is the task of assigning one of many class labels to a short video clip containing an action. Long untrimmed videos further carry a burden of multi-label assignment.
Action recognition in videos plays a crucial role in many applications, e.g. visual surveillance~\cite{lin2008human}, sport video analysis~\cite{soomro2014action}, human-machine interaction~\cite{vantigodi2013real}, video object tracking~\cite{wang2016temporal}, etc. It has piqued the interest of the computer vision and deep learning communities, owing to the fact that the performances of state-of-the-art approaches are still well below human-level performance. Action recognition is a more complicated task when compared to still image classification because the temporal domain introduces variations in motion and viewpoints which have to be accounted for. Additionally, the use of a moving camera rather than a static camera introduces variations that could make optical flow based features less reliable. Besides, the interactions of multiple objects in actions make the classification task even more challenging.

Intuitively, action recognition requires models capable of learning key features such as:

\textbf{Appearance features:} Some actions are defined by certain special objects. For the ``Blowing Candles'' class in the UCF-101 dataset~\cite{soomro2012dataset}, the presence of a candle in any one frame is sufficient to correctly classify the action.

\textbf{Short-term motion features:} Some actions are characterized by particular short-term motion with large variations of appearance. For the ``Boxing Speed Bag'' class in the UCF-101 dataset, a short optical flow snippet of the video clip is sufficient to correctly classify the action.

\textbf{Long-term trajectory features:} Similarly, some actions depend on long-term object trajectories which are defined by appearance and motion jointly. For the ``Golf Swing'' class in the UCF-101 dataset, a longer optical flow snippet of the video clip composed of the long-term motion of the arm and golf club is required to correctly classify the action.

\textbf{Object interaction features:} Multi-object based actions are frequently observed and entails modeling of object interactions.  For the ``Frisbee Catch'' class in the UCF-101 dataset, the interaction of the Frisbee moving between 2 persons is required to correctly classify the action. 
Note that this category of features can be further divided into object interactions \textit{across space} and \textit{across time}, but exploring specific scenarios where they are applicable in an isolated fashion becomes harder.

Clearly, key features stated above do not present themselves in a mutually exclusive form in popular benchmark datasets; appearance features would be useful regardless of whether the action possessed interactions between objects. Nevertheless, one popular approach that has brought recent success to modeling these features is that of signal decomposition. Inspired by the discovery that the Human Visual System has separate processing pathways for different types of signals such as fast and slow motion~\cite{kaiser1996human}, researchers have employed multi-stream architectures to process each of these features separately. This paper is also motivated by the idea of signal decomposition. %although we build on this idea, we do not explore this connection further.

% the description in this paragraph and in related work section for two-stream is kind of duplicate
Most significantly, the two-stream architecture of Simoyan and Zisserman \cite{simonyan2014two} pioneered research in this area for action recognition from videos. They design two parallel 2D ConvNet streams to learn appearance features from RGB images and motion features from optical flow fields. An alternative to the optical flow stream also presented itself in the form of 3D convolutions \cite{ji20133d}, although they have been used together too since then \cite{carreira2017quo}. Since 2D convolution on an optical flow stream and 3D convolution were deemed to not be able to capture sequential information required to model long-term trajectory features well,  researchers turned to explore ways to address this next. The use of LSTMs following 2D convolutions \cite{ng2015beyond, li2018videolstm} and various temporal fusion strategies \cite{cherian2017generalized, diba2017deep, kar2017adascan, bilen2017action} emerged as competitive candidates. However, object interaction features were left unexplored for the most part until recently where researchers used a self-attention mechanism to develop non-local neural networks \cite{wang2018non}. Their use of a ``non-local operation'' is able to model long-range interactions between objects in space and time, where high-level latent feature vectors built on top of a series of convolution blocks represent the objects. But their network is applied to short snippets cropped from the original videos. As such, they fail to explore truly modeling long-term trajectory features from information across the full lengths of videos.

In this paper, we explore the novel idea of introducing the non-local operation from \cite{wang2018non} into an LSTM module to create a Relational LSTM. We hypothesize that introduction of our ``relational LSTM'' block into a two-stream architecture would aid in the modeling of features that capture object interactions, while retaining the property of LSTM to model long-term trajectory information. Our main contributions can be summarized as follows:

\begin{itemize}
    \item We develop a novel Relational-LSTM module that models object interaction features and can be inserted into existing diverse architectures as a plug-in module.
    \item We incorporate the Relational-LSTM module into a two-stream, two-branch architecture to perform video action recognition.
    \item We show experimentally through ablation studies that the introduction of our module leads to clear and unquestionable gains in performance.
    \item Our architecture should be considered the new state-of-the-art for video action recognition among LSTM-based architectures, beating the current best architecture by 1.2\% on UCF-101 and 5.2\% on HMDB-51.
    \item Our architecture performs comparably to the top-tier of state-of-the-art architectures overall on UCF-101 and HBDM-51 datasets.
    \item Our architecture outperforms state-of-the-art methods with comparable schema on the large-scale Charades dataset by 3.2\%.
\end{itemize}

The rest of the paper is organized as follows. In section \ref{sec:related work}, we discuss related work. Section \ref{sec:method} describes our Relational LSTM module. Experimental results and their analysis are presented in Section \ref{sec:exp}. Finally, the conclusions and potential for future work are discussed in Section \ref{sec:conclusion}.

\section{Related Work}
\label{sec:related work}
\subsection{Deep learning for appearance and short-term motion features}
The success of 2D ConvNets did not immediately follow for video tasks, where hand-crafted features of Improved Dense Trajectories dominated. It was not until the work of \cite{simonyan2014two} that deep learning approaches started to show comparable performance. 

\textbf{Two-stream ConvNets}: 
In \cite{simonyan2014two}, the authors employ a two-stream architecture with 2D ConvNets to learn appearance and motion features to aid classification. They show that 2D ConvNets are by themselves capable of capturing short-term motion features with densely stacked optical flow fields as inputs. They average the predictions from a single RGB image and a stack of 10 consecutive optical flow fields after feeding them through two separate 2D ConvNets with identical structure. Based on this two-stream architecture, Wang $et$ $al.$~\cite{wang2016temporal} propose the averaging pooling operator to aggregate multiple frame-level predictions into a video-level prediction to model long-range temporal structure over the entire video. %Two-stream fusion methods have also been proposed. 
% \citeauthor{jain2017fusionseg} \shortcite{jain2017fusionseg} explore an approach to perform late fusion of the spatial and temporal streams by picking the maximum from the prediction scores of the streams or the multiplication of the two prediction scores. 
% \citeauthor{feichtenhofer2016spatiotemporal} \shortcite{feichtenhofer2016spatiotemporal} exploit an early fusion approach, which allows learning of spatiotemporal features via early interaction of the spatial and temporal streams.

\textbf{3D ConvNets}: When viewing a video as a sequential stack of RGB images, it is natural to think of extending 2D convolution to the temporal dimension to model spatio-temporal features in videos. In early stages, Ji $et$ $al.$~\cite{ji20133d} have attempted to replace pre-computed complex hand-crafted features with 3D ConvNets, but their network is still quite shallow with only three convolutional layers. Following their work, Tran $et$ $al.$~\cite{tran2015learning} further exploit 3D ConvNets' properties under various video datasets
% or exploit a generalized version of 3D ConvNets applying to various training datasets.
and experimentally show that 3D ConvNets are competent for learning appearance and short-motion features. Inflated 3D ConvNet (I3D) proposed by Carreira and Zisserman~\cite{carreira2017quo} makes full use of successful pre-trained image classification architectures by inflating all the filters and pooling kernels from 2D to 3D and achieves state of the art performance on UCF-101~\cite{soomro2012dataset} and HMDB-51~\cite{kuehne2011hmdb} datasets. Their competitive performance drove research in this direction. 
The authors in \cite{xie2017rethinking} show that factorizing 3D convolution of I3D into a 2D spatial convolution and a 1D temporal convolution, analogous to spatial factorization in Inception-v2 \cite{szegedy2016rethinking}, yields slightly better accuracy. 

\subsection{Sequence modeling for long-term trajectory features}
The aforementioned methods successfully model appearance and short-term motion features. However, with regard to long-term trajectory information, they either used unsuitable inputs (frames not spanning long temporal range), or they used the 3D convolution operation or optical flow inputs, which capture only \textit{local} temporal properties. Alternatively, diverse methods have attempted to encode \textit{long-term} trajectory information. Some authors~\cite{wang2011action, wang2013action, peng2014action} make use of dense point trajectories by tracking densely sampled points using optical flow fields. Subsequently, this hand-crafted shallow video representation was replaced by deep representations learned from neural networks. The basic idea is to sequentially aggregate frame-level feature representations extracted from either 2D ConvNets or 3D ConvNets so that long-term trajectory information is encoded into the deep video-level representations. This could be done using either recurrent neural networks (RNNs) such as LSTMs or temporal feature fusion methods such as temporal pooling. Other methods to aggregate temporal information include the use of long-term feature banks \cite{wu2019long}, multi-scale temporal convolutions \cite{hussein2019timeception} and grammar models \cite{qi2018generalized}.

\textbf{RNN-based architectures}: The sequential modeling ability of LSTMs makes them appealing to use for capturing long-range temporal dynamics in videos. In \cite{baccouche2011sequential}, the authors propose applying an LSTM to high-level feature vectors extracted from 3D ConvNets, but they only apply it on short video snippets of 9 frames. Ng $et$ $al.$ \cite{ng2015beyond} add five stacked LSTM layers before the last fully connected layer of the two-stream ConvNets \cite{simonyan2014two} and slightly improve the performance. Wang $et$ $al.$ \cite{wang2016hierarchical} design a hierarchical attention network, which is implemented by skipping time steps in higher layers of the stacked LSTM layers. Additionally, the authors in \cite{wang2019revisiting} employ ConvLSTM along with an attention mechanism \cite{xu2015show} to automatically focus on sequential saliency regions from high-level appearance feature maps.

\textbf{Temporal feature fusion architectures}: 
Temporal feature pooling~\cite{ng2015beyond, bilen2017action} is the most popular temporal feature fusion method, which usually uses either max-pooling or average-pooling over the temporal dimension to aggregate frame-level features. Furthermore, the authors \cite{kar2017adascan, bilen2017action} propose adaptive temporal feature pooling by simultaneously learning an importance score for each frame and use it as weight in the pooling process. Cherian $et$ $al.$ \cite{cherian2017generalized} propose generalized ranking pooling, which projects all frame-level features together into a low-dimensional subspace and use an SVM classifier on the subspace representation. The subspace is parameterized by several orthonormal hyperplanes and is designed to have a property of preserving the temporal order of video frames. Besides temporal pooling, Diba $et$ $al.$ \cite{diba2017deep} introduce a temporal linear encoding method, where they first aggregate frame-level features using element-wise multiplication and then project it to a lower dimensional feature space using bilinear model.

\subsection{Non-local operation for object interaction features}
Both the convolution and recurrent operations compute spatial and temporal features respectively in a primarily \textit{local} fashion. Long-range dependencies are then modeled through applying these local operations repeatedly, often accompanied by downsampling, to propagate signals across space and time domains. The non-local operation \cite{wang2018non} is one hypothesized solution to handle the remaining problem of object interaction modeling.

\textbf{Relation reasoning}
Exploring object interactions is equivalent to reason the relations of objects. Recently, the authors in \cite{santoro2017simple} propose Relation Network (RN), which is a neural network module primarily for relation reasoning. It uses one MLP layer on top of a batch of feature vector pairs to learn pairwise relations, where each instance in the batch is a pair of feature vectors at two particular positions in the input feature maps. They use this RN module on the visual question-and-answer problem and achieve super-human level performance. Following their work, Zhou $et$ $al.$\cite{zhou2017temporal} explore its usage on temporal relation reasoning in videos, which is implemented by sparsely sampling frames from videos and employing RN module to learn the causal relations among frames. A similar work for exploring object relations is proposed by \cite{hu2018relation}, where they use 'Scaled Dot-Product Attention'~\cite{vaswani2017attention} to compute object relations. Another way to learn object interactions is using Graph Parsing Neural Network (GPNN)~\cite{qi2018learning}, where a graph is built on latent features and the edge weights of the graph are learnt through training by message passing.

\textbf{Non-local operation} 
The non-local operation can be considered as a general form of 'Scaled Dot-Product Attention', as mentioned in \cite{wang2018non}. The key idea of non-local operation is that the output features of a position are computed as a weighted sum of the features from all positions in the input feature maps, which allows contributions from features in distant positions. The non-local idea originates from \cite{buades2005non} for image denoising, where the estimated value in pixel $i$ is computed as the weighted average of all the pixels in the image. In \cite{wang2018non}, they leverage it to design a non-local block for a neural network, which can be used as a plug-in module inserted into diverse neural network architectures.

\section{Relational LSTM Module}
\label{sec:method}
Considering the importance of object interaction features in videos, we propose a Relational LSTM module, which not only inherits the sequential modeling ability from LSTM but also incorporates spatial relation reasoning and temporal relation reasoning through a non-local operation. More specifically, we generalize the non-local operation in \cite{wang2018non} to compute spatial relations among input features at a single snippet, and to compute temporal relations between input features and past learned features at previous time steps. Meanwhile, because of the use of LSTM, we create video-level feature representations by using selected snippets from the whole video, which inherently encodes long-term trajectory information in our representations. As is common in the deep learning community, the objects we refer to in our method are instance-level, which are represented by high-level feature vectors built on top of a series of convolution blocks.

\textbf{Non-local operation}: We first review the non-local operation defined in \cite{wang2018non}. Given input feature maps $\boldsymbol{X} \in \mathbb{R}^{N \times C}$, where $N$ represents the number of positions in $\boldsymbol{X}$, and $C$ represents the number of dimensions of feature vector at each position. If we represent $\boldsymbol{X}$ as $\{\boldsymbol{x}_i\}_{i=1}^{N}$, the output $\boldsymbol{z}_i$ at $i$-th position of response feature maps  $\boldsymbol{Z} \in \mathbb{R}^{N \times C}$ is a weighted sum over all input feature vectors, as shown in Equation \textbf{{\ref{eq:non-local-operation}}}.

\begin{equation}
\label{eq:non-local-operation}
\begin{aligned}
\boldsymbol{z}_i &= \sum_{j=1}^{N}\omega_{ij}g(\boldsymbol{x}_j) 
\\
\omega_{ij} &= \frac{f(\boldsymbol{x}_i,\boldsymbol{x}_j)}{\mathcal{C}(\boldsymbol{X})}
\end{aligned}
\end{equation}
where $i, j, k \in \mathbb{R}^N$ are position indices, $f(\boldsymbol{x}_i,\boldsymbol{x}_k)$ represents compatibility function computing the similarity between $\boldsymbol{x}_i$ and $\boldsymbol{x}_j$, and $g(\boldsymbol{x}_j)$ is a unary function computing a representation of the input feature vector at $j$-th position. $\mathcal{C}(\boldsymbol{X})$ is the normalization factor, usually denoted as $\mathcal{C}(\boldsymbol{X}) = {\sum_{k=1}^{N} f(\boldsymbol{x}_i,\boldsymbol{x}_k)}$ or $\mathcal{C}(\boldsymbol{X}) = N$. 

The non-local operation in Equation \textbf{{\ref{eq:non-local-operation}}} considers all positions in $\boldsymbol{X}$ regardless of positional distance, which is in sharp contrast to a convolutional operation of considering absolutely local neighborhood. This non-local operation explores the relations of features globally, which can be adopted to learn object interaction features far from each other in a spatio-temporal layout. 

\textbf{Generalized non-local operation}: First of all, we generalize the non-local operation to compute relations between any two feature maps $\boldsymbol{X}\in \mathbb{R}^{N \times C_X}, \boldsymbol{Y}\in \mathbb{R}^{N \times C_Y}$. Given feature maps $\boldsymbol{Y}$, we compute the output $\boldsymbol{Z} \in \mathbb{R}^{N \times C_Z}$ with respect to $\boldsymbol{X}$ as shown in Eq.~\textbf{{\ref{eq:non-local-operation-general}}}
\begin{equation}\label{eq:non-local-operation-general}
\begin{aligned}
\boldsymbol{z}_i &= \sum_{j=1}^{N}\omega_{ij}g(\boldsymbol{y}_j) 
\\
\omega_{ij} &= \frac{f(\boldsymbol{x}_i,\boldsymbol{y}_j)}{\sum_{k=1}^{N} f(\boldsymbol{x}_i,\boldsymbol{y}_k)}
\end{aligned}
\end{equation}
This general form of the non-local operation can also be interpreted using the attention mechanism in \cite{vaswani2017attention}. Given a query feature vector $\boldsymbol{x}_i$ and a set of input feature vectors $\{\boldsymbol{y}_i\}_{i=1}^{N}$, the output $\boldsymbol{z}_i$ is computed as an attentional weighted sum of all input feature vectors, where the attentional weights are computed by a compatibility function of the query feature vector and input feature vectors. In the rest of the paper, we abbreviate this general form as $\boldsymbol{Z} = r(\boldsymbol{X}, \boldsymbol{Y})$.

\textbf{Relational LSTM}: Given a video $V$, we first divide it into $T$ segments \{$S_1, S_2, ..., S_T$\} of equal duration, and randomly sample one short snippet $K_t$ from its corresponding segment $S_t$. Suppose $\boldsymbol{X}_t \in \mathbb{R}^{H \times W \times C}$ for $t = 1, ..., T$ represents the high-level feature maps obtained after feeding $K_t$ through some Convolution layers of a pre-trained CNN model, where $C$ is the number of feature maps, and $H$ and $W$ are the spatial height and width of each feature map. After extracting $\{X_1, ..., X_T\}$ from a convolution layer, a temporal aggregation function is required to encode these snippet-level feature maps into video-level feature maps. Inspired by the deep learning architecture LSTMs, which not only inherit the sequential modeling ability from vanilla RNNs but can also capture long-term dependencies through the memory cell mechanism, we employ LSTM-based architectures to this end. 

Usually, given feature maps $X_t$ which preserve spatial layout, ConvLSTM~\cite{xingjian2015convolutional, wang2019learning} or Bidirectional ConvLSTM~\cite{song2018pyramid} are the natural choices as the aggregation function because it encodes spatial information through its convolutional operations. However, the experimental results in \cite{sun2017lattice} have shown that ConvLSTM did not perform well on this recognition task, and it could not capture crucial object interaction features because of its convolution operations, which are applied to a \textit{local} receptive field. We introduce the generalized non-local operation into LSTM architecture and present a new module called Relational LSTM. 
It differs from regular LSTM and ConvLSTM~\cite{xingjian2015convolutional} in the aspect that the general form of non-local operation is used in both input-to-state transitions and state-to-state transitions. The key equations of Relational LSTM are shown in Equation \ref{rl:relational_lstm}.

\begin{equation}
\label{rl:relational_lstm}
\begin{aligned}
\boldsymbol{i}_t &=  \sigma[r_{ix}(\boldsymbol{X}_t, \boldsymbol{X}_t) + r_{ih}(\boldsymbol{X}_t, \boldsymbol{H}_{t-1})] \\
\boldsymbol{f}_t &=  \sigma[r_{fx}(\boldsymbol{X}_t, \boldsymbol{X}_t) + r_{fh}(\boldsymbol{X}_t, \boldsymbol{H}_{t-1})] \\
\boldsymbol{o}_t &=  \sigma[r_{ox}(\boldsymbol{X}_t, \boldsymbol{X}_t) + r_{oh}(\boldsymbol{X}_t, \boldsymbol{H}_{t-1})] \\
\boldsymbol{g}_t &=  \tanh[r_{gx}(\boldsymbol{X}_t, \boldsymbol{X}_t) + r_{gh}(\boldsymbol{X}_t, \boldsymbol{H}_{t-1})] \\
\boldsymbol{C}_t &= \boldsymbol{f}_t \circ \boldsymbol{C}_{t-1} + \boldsymbol{i}_t \circ \boldsymbol{g}_t \\
\boldsymbol{H}_t &= \boldsymbol{o}_t \circ \tanh(\boldsymbol{C}_t) 
\end{aligned}
\end{equation}

Here inputs $\boldsymbol{X}_t$, memory cell $\boldsymbol{C}_t$, hidden state $\boldsymbol{H}_t$, and gates $\boldsymbol{i}_t, \boldsymbol{f}_t,\boldsymbol{o}_t, \boldsymbol{g}_t$ have same functionalities as traditional LSTM. $\sigma$ represents the logistic sigmoid non-linear activation function and $\tanh$ represents the  hyperbolic tangent non-linear activation function.

%%%%%%%%%%%%%%%%%%%%%%%%%%%
\begin{figure}[htb]
\centering
\mbox{
    \includegraphics[height=3.3in, width=3.0in]{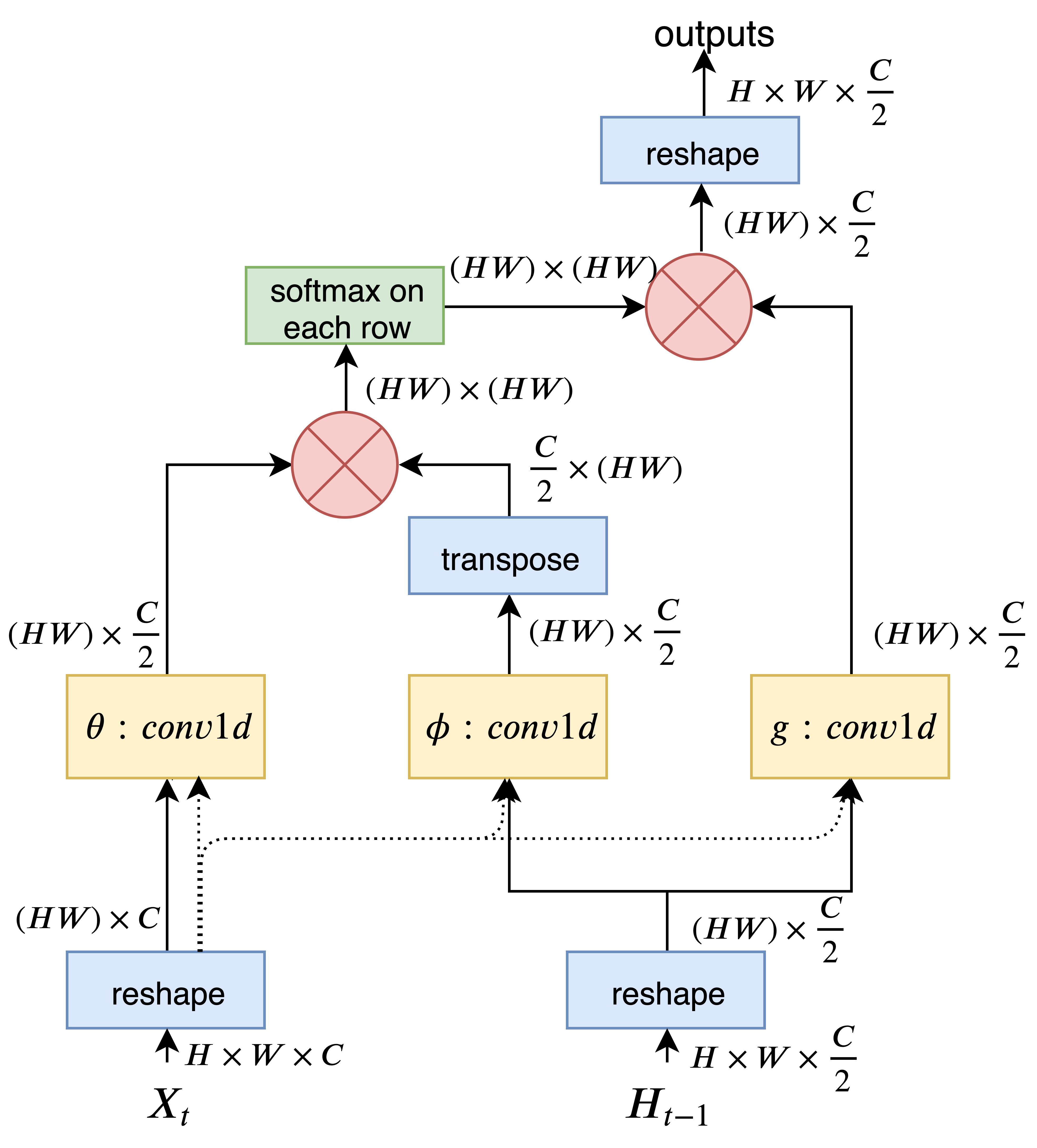}
  }
   \caption{\textbf{Generalized non-local operations $r(\boldsymbol{X}_t,\boldsymbol{X}_t)$ and $r(\boldsymbol{X}_t, \boldsymbol{H}_{t-1})$ in Relational LSTM.} ``$conv1d$'' denotes 1D convolutional operation, ``$\otimes$'' denotes matrix multiplication. Because the only difference between $r(\boldsymbol{X}_t,\boldsymbol{X}_t)$ and $r(\boldsymbol{X}_t, \boldsymbol{H}_{t-1})$ is the inputs, we use dashed arrow for inputs of $r(\boldsymbol{X}_t,\boldsymbol{X}_t)$, and use solid arrow for inputs of $r(\boldsymbol{X}_t, \boldsymbol{H}_{t-1})$.}
\label{fig:generalized_non_local}
\end{figure}
%%%%%%%%%%%%%%%%%%

We disentangle the mixed spatial-temporal relational reasoning. To ensure spatial relational reasoning is used, we adopt $r(\boldsymbol{X}_t, \boldsymbol{X}_t)$ in input-to-state transitions to model the feature interactions in same feature maps regardless of their relative positional distance. Regarding temporal relational reasoning, to model feature interactions of $X_t$ with  $H_{t-1}$ which stores all important information from preceding feature maps, we adopt $r(\boldsymbol{X}_t, \boldsymbol{H}_{t-1})$ in state-to-state transitions. The detailed implementations of $r(\boldsymbol{X}_t, \boldsymbol{X}_t)$ and $r(\boldsymbol{X}_t, \boldsymbol{H}_{t-1})$ are shown in Fig.~\ref{fig:generalized_non_local}. In our implementations, we adopt the shape of $\boldsymbol{X}_t$ as $H \times W \times C$, and set $\boldsymbol{H}_{t-1}$ as $H \times W \times \frac{C}{2}$ to reduce memory cost. Given $\boldsymbol{X}_t$ and $\boldsymbol{H}_{t-1}$ as inputs of $r(\boldsymbol{X}_t, \boldsymbol{H}_{t-1})$, we first reshape $\boldsymbol{X}_t$ to $(HW)\times C$ and $\boldsymbol{H}_{t-1}$ to $(HW)\times \frac{C}{2}$. Then we apply the generalized non-local operation defined in Eq. \textbf{{\ref{eq:non-local-operation-general}}} on them. We employ Embedded Gaussian function as $f(\boldsymbol{x}_i, \boldsymbol{y}_j)$ (shown in Equation \ref{eq:f(x,y)}) :
\begin{equation}
\label{eq:f(x,y)}
f(\boldsymbol{x}_i, \boldsymbol{y}_j) = e^{\theta(\boldsymbol{x}_i)^T \phi(\boldsymbol{y}_j)}
\end{equation}
where $\theta(\boldsymbol{x}_i) = W_{\theta}x_i$ and $\phi(\boldsymbol{y}_j) = W_{\phi}y_j$ are two linear embedding function (implemented by 1D convolutional layer in Fig.~\ref{fig:generalized_non_local}). We consider $g(\boldsymbol{y}_j)$ also in the form of a linear embedding function expressed as $g(\boldsymbol{y}_j) = W_g\boldsymbol{y}_j$. The normalization function in the generalized non-local operation is implemented as a softmax layer. It is worth noting that even though we flatten the spatial layout when feeding $\boldsymbol{X}_t$ and $\boldsymbol{H}_{t-1}$ to the Relational LSTM block, we still preserve their relative positional information through Relational LSTM block and obtain output hidden feature maps $H_t$ of shape as $(HW)\times \frac{C}{2}$ , so that we can reshape $H_t$ back to a shape of $H \times W \times \frac{C}{2}$. As for $r(\boldsymbol{X}_t, \boldsymbol{X}_t)$, it is implemented in the same way as mentioned above except that the inputs to it are both $\boldsymbol{X}_t$ (presented as dashed arrow in Fig~\ref{fig:generalized_non_local}).

%%%%%%%%%%%%%%%%%%%%%%%%%%%
\begin{figure*}[htb]
\centering
\mbox{
    \includegraphics[height=2.8in, width=\linewidth]{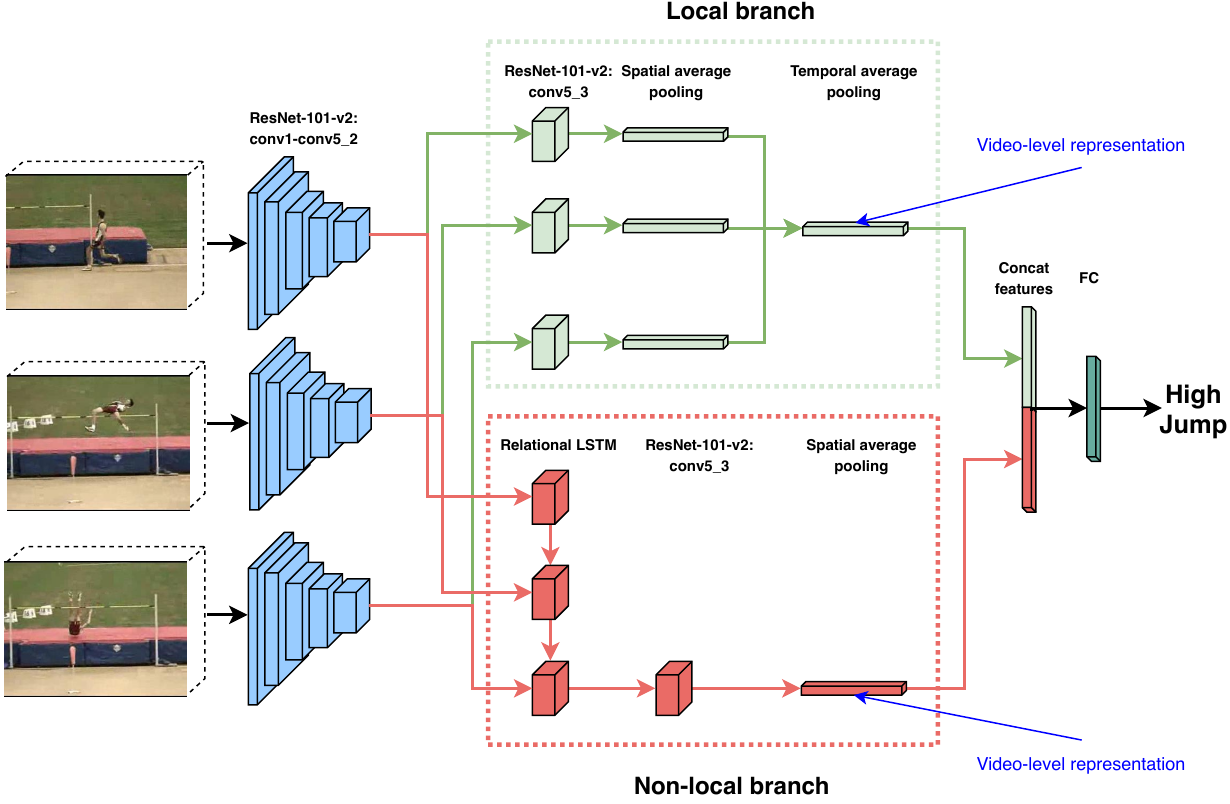}
  }
   \caption{\textbf{Network architecture.} Only spatial stream is shown; the temporal stream is identical in structure. Given an input video, we divide it into T segments (T=3 in this case for succinct illustration) of equal duration, and randomly sample one short snippet from its corresponding segment. The short snippet is either a single RGB image (spatial stream) or a sequence of 10 optical flow fields (temporal stream). After feeding the selected short snippets through ResNet-101-v2 conv1 layer to ResNet-101-v2 conv5\_2 layer, we have two separate branches. The local branch (green dotted box) captures local appearance and short-term motion features from snippets, and generates a video-level feature representation using temporal average pooling. The non-local branch (red dotted box) is for relation reasoning using Relation LSTM module. Eventually, we obtain an overall video-level feature representation by concatenating feature vectors generated by the two branches, and add one Fully Connected layer and optimize using a standard softmax with cross entropy classification loss.}
\label{fig:network_architecture}
\end{figure*}
%%%%%%%%%%%%%%%%%%

\section{Network Architecture}
\label{sec:architecture}
We build our architecture on top of ``two-stream ConvNets'' \cite{simonyan2014two}, where a spatial stream operates on RGB images and a temporal stream operates on sequences of 10 optical flow fields, and their prediction scores are fused by weighted averaging. Our network architecture of each stream is shown in Fig. \ref{fig:network_architecture}. Obviously, the backbone network for image-level feature extraction plays an important role in our architecture and our architecture is compatible with most deep feature extractor networks, e.g. VGG-16~\cite{Simonyan15}, ResNet~\cite{he2016deep}, BN-Inception~\cite{li2018videolstm}, etc. However, our goal here is to evaluate the effectiveness of our proposed Relational LSTM module and two-branch network architecture, we fix the backbone network consistently through all our experiments. The ResNet-101-v2~\cite{he2016identity} has been employed as the backbone network of our architecture. The detailed building blocks of ResNet-101-v2 are shown in Table~\ref{table:ResNet-v2}. We extract the feature maps $\{X_1, ..., X_T\}$ after ResNet-101-v2 conv5\_2 layer, and split our network into two branches. Two primary factors are determining our decision of the place for extracting the feature maps: 1. obtain relatively high-level features to represent objects; 2. have some convolutional block after the Relational LSTM module to further explore relative positional information.

\begin{table}[htb]
\begin{center}
\begin{tabular}{c|c|c}
\hline
layer name & 101-layer & output size \\ 
\hline
conv1 & $7 \times 7$, 64, stride 2,2 & $112 \times 112$\\
\hline
pool1 & $3 \times 3$ max, stride 2,2 & $ 56 \times 56$\\
\hline
conv2\_x & $\begin{bmatrix} 
1 \times 1, 64 \\
3 \times 3, 64 \\
1 \times 1, 256
\end{bmatrix}$ $\times 3$ & $56 \times 56$ \\
\hline
conv3\_x & $\begin{bmatrix} 
1 \times 1, 128 \\
3 \times 3, 128 \\
1 \times 1, 512
\end{bmatrix}$ $\times 4$ & $28 \times 28$\\
\hline
conv4\_x & $\begin{bmatrix} 
1 \times 1, 256 \\
3 \times 3, 256 \\
1 \times 1, 1024
\end{bmatrix}$ $\times 23$ & $14 \times 14$\\
\hline
conv5\_x & $\begin{bmatrix} 
1 \times 1, 512 \\
3 \times 3, 512 \\
1 \times 1, 2048
\end{bmatrix}$ $\times 3$ & $7 \times 7$\\
\hline
\multicolumn{2}{c|}{global average pool, fc, softmax} & $1 \times 1$ \\
\hline
\end{tabular}
\end{center}
\caption{\textbf{ResNet-101-v2 architecture.} Residual blocks are shown in brackets. The input is $224 \times 224 \times 3$, and downsampling is performed at conv3\_1, conv4\_1, conv5\_1 with a stride of 2, 2.} 
\label{table:ResNet-v2}
\end{table}

\textbf{Local branch}: We design this branch for learning information that can be learned via local operations. The information can either be appearance features with RGB images as inputs, or short-motion features with optical flow fields as inputs. Specifically, we continue adopting ResNet-101-v2 architecture (conv5\_3, global spatial average pooling) on every individual short snippet to obtain snippet-level feature representations, then integrating them to a video-level representation by temporal average pooling. 

\textbf{Non-local branch}: We design this branch for learning object interactions in space and time, taking into account long-range temporal dependencies. The Relational LSTM module in this branch can not only perform spatial relation reasoning and temporal relation reasoning among those snippets, but also obtain a video-level feature representation natively. As the spatial layout is preserved through the Relational LSTM module, and we further explore relative positional information by adding one residual block (ResNet-101-v2 conv5\_3 layer) after Relational LSTM module. Specifically, in our design of Relational LSTM module, we add one batch normalization (BN) layer just before the first reshaping operator, and one $1 \times 1$ convolutional layer after the outputs to increase the number of feature maps from $H \times W \times \frac{C}{2}$ to $H \times W \times C$. We initialize $H_0$ of the Relational LSTM module with zeros, assuming that no information has been observed when the video starts. 

Finally, we concatenate the video-level representations generated by the two branches as a complement to each other, and add one fully connected layer to perform classification by taking the argmax. 

\section{Experiments}
\label{sec:exp}
In this section, we first introduce the action recognition datasets we conduct experiments on and implementation details of our architecture including training and inferencing. Then, we study different aspects of our network on split 1 of UCF-101 dataset. Finally, we compare our architecture with state-of-the-art methods. 

\subsection{Datasets}
We first conduct a series of experiments on two challenging video action recognition benchmark datasets, UCF-101~\cite{soomro2012dataset} and HMDB-51~\cite{kuehne2011hmdb}. The UCF-101 dataset consists of 13,320 short video clips with 101 action classes. The HMDB-51 dataset consists of 6,766 short video clips with 51 action classes. Both datasets contain video clips of typically 5-10 seconds and have more than 100 video clips in every action class. For both datasets, we use the provided evaluation schema and report the mean average accuracy over 3 training/testing splits. We also evaluate our architecture on the more complex and large-scale Charades dataset~\cite{sigurdsson2016hollywood}. The Charades dataset contains 9,848 videos of daily activities with 157 action classes. Unlike UCF-101 and HMDB-51 datasets, Charades dataset has longer video durations (around 30 seconds on average) and more than one action is performed concurrently or sequentially in each video. For evaluation on Charades, a single video is assigned to multiple class labels, and the standard mean avarage precision (mAP) measure is employed as the evaluation metric.

\subsection{Implementation details}
\label{sec:implementation-details}
\textbf{Training}: We separately train the two streams of our architecture. The backbone ResNet-101-v2 is pre-trained on ImageNet. We employ the same data augmentations used in \cite{wang2016temporal} e.g. horizontal flipping and multi-scale cropping. We adopt mini-batch stochastic gradient descent with a momentum of 0.9 and weight decay of 0.0005 as our optimizer. And we add a dropout layer right before the final fully connected layer. We train our model with Batch Normalization~\cite{ioffe2015batch}. Traditionally, batch mean and variance are adopted for training in Batch Normalization, and moving mean and moving variance are adopted for inferencing. However, because of the limited number of training instances and batch size, the estimates of mean and variance from each batch is highly variant from moving mean and moving average. It leads to severe over-fitting due to divergent distributions of data in training and inferencing stages after BN layer, as mentioned in~\cite{wang2016temporal}. Although completely freezing mean and variance parameters of all BN layers work well in practice, we found that adopting moving mean and moving variance in the training stage with a very small momentum (e.g. 0.001, 0.0005) to gradually update them makes more sense as they are progressively learning population mean and variance of the training dataset.

For the spatial stream, we initialize parameters for conv1-conv5\_2 layers and conv5\_3 layer in the local branch from pre-trained ResNet-101-v2 model, and initialize Relational LSTM module and conv5\_3 layer in the non-local branch using Xavier-Glorot initialization~\cite{glorot2010understanding}. We set dropout rate to 0.8 and batch size to 24 (when T = 8). We train the model for 50 epochs. The learning rate starts at $0.0005$ and is reduced by a factor of 10 after 35-th epoch and 45-th epoch. 

For the temporal stream, we employ the cross-modality initialization strategy used by \cite{wang2016temporal}, where the parameters are initialized from our trained spatial stream model. We set dropout rate to 0.7 and batch size to 24 (when T=8). We train the model for 60 epochs. The learning rate starts at $0.0005$ and is reduced by a factor of 10 after 45-th epoch and 55-th epoch. 

\textbf{Inference}: For testing our architecture, we sample 1 RGB image or a sequence of 10 optical flow fields from the same position of each segment ($T=8$ segments by default) to form one testing group. Meanwhile, we generate 5 crops (4 corners and 1 center) of $224 \times 224$ from the sampled images in the group. And we generate 4 groups by sampling from 4 positions with equal temporal spacing. So the final prediction scores for each stream are obtained by averaging over the 20 testing examples.  We fuse the prediction scores of the two streams by weight averaging and the fusion is conducted before softmax normalization. Our empirical experiments suggest that the weight should be chosen around 0.5 for each stream in general.

\subsection{Experimental evaluation on Relational LSTM network}
In this experiment, we investigate how well standalone Relational LSTM module can perform on this task. We exclude local branch from our architecture and name the rest Relational LSTM network, and conduct experiments on it using split 1 of UCF-101 dataset. The spatial stream is initialized from pre-trained ResNet-101-v2 on ImageNet, and the temporal stream is initialized from the temporal stream of Temporal Segment Networks (TSN) proposed in \cite{wang2016temporal}. We compare its performance with our implementation of TSN using ResNet-101-v2 as backbone in Table \ref{table:r-lstm vs tsn}. From the table, we observe that Relational LSTM network itself achieves a similar overall performance as TSN. Combining the two models yields obvious increases in both spatial stream and temporal streams, indicating that TSN and our Relational LSTM network are complementary to some extent. This complementary behavior is to be expected as the Relational LSTM network is designed to focus more on object interaction and long-term motion features, whereas TSN emphasizes learning appearance and short-motion features. Inspired by this complementary behavior, we designed the two-branch network architecture, where local branch aims to explore more on local features (e.g., appearance and short-motion features) by spatial-temporal pooling operations, and non-local branch aims to explore more on non-local features (e.g., object interactions and long-term motion features) through our proposed Relational LSTM module. It is worth noting that our local branch captures appearance/short-motion features in a different way compared to TSN. Our local branch aggregates snippet-level feature representation into video-level representation before FC layer, whereas TSN aggregates multiple snippet-level predictions after FC layer into a video-level predictions.

\begin{table}[htb]
\begin{center}
\begin{tabular}{|c|c|c|c|}
\hline
Methods & Spatial & Temporal & Two-stream \\
\hline
TSN & 87.0\% & 88.5\% & 93.8\%\\
\hline
R-LSTM & 86.9\% & 87.5\% & 93.8\%\\
\hline
TSN + R-LSTM & 88.6\% & 89.1\% & 94.2\% \\
\hline
\end{tabular}
\end{center}
\caption{Performance of Relational LSTM network, TSN and the ensemble of Relational LSTM network and TSN on UCF-101 (split 1). Relational LSTM network uses 8 segments ($T=8$) in this experiment. We choose the best fusion weights in late fusion, 0.5 for spatial stream in Relation LSTM network, and 0.35 for spatial stream in TSN, and average weight for each stream in the ensemble.} 
\label{table:r-lstm vs tsn}
\end{table}

\subsection{Ablation studies}
In this section, we explore different aspects of our architecture. The experiments are all conducted on split 1 of UCF-101 dataset. 

\textbf{The number of input video segments}: The most important parameter influencing our model performance is the number of input video segments. We mutate our architecture with a different number of input video segments from $T=6$ to $T=12$. The results are shown in Table \ref{table:diff-segments}. From the table, we see that there is a large increase ($1.5\%$) in the spatial stream from $T=6$ to $T=8$, which implies that more segments in a video can provide richer information serving as a better video representation. However, on continuing to increase the number of segments, the performance takes a hit. This could partly owe to the na\"ive temporal pooling operation in the local branch causing a damping of the correct action signal. Therefore, we set $T=8$ for the rest of our experiments.

\begin{table}[htb]
\begin{center}
\begin{tabular}{|c|c|c|c|}
\hline
Number of segments & Spatial & Temporal & Two-stream \\
\hline
6& 87.2\% & 87.9\% & 94.2\% \\
\hline
8 & \textbf{88.7\%} & \textbf{88.1\%} & \textbf{94.4\%} \\
\hline
10 & 88.1\% & 87.5\% & 94.1\% \\
\hline
12 & 87.4\% & 87.4\% & 93.5\% \\
\hline
\end{tabular}
\end{center}
\caption{Performance of our architecture with different number of input video segments on UCF-101 (split 1).} 
\label{table:diff-segments}
\end{table}

\textbf{Effect of introducing non-local branch}: Since there are two branches in our architecture, we were compelled to quantify the benefit of introducing each branch into our architecture. In this experiment, we compare the performance of excluding non-local branch vs. excluding local branch vs. including both branches in our architecture. The results are shown in Table \ref{table:local-two}, and we find that there are improvements in each of the individual streams as well as overall. These small improvements on a large-scale dataset such as UCF-101 indicate that the contributions of the non-local branch in our architecture are significant and crucial. 

\begin{table}[htb]
\centering
\begin{tabular}{|c|c|c|c|}
\hline
Methods & Spatial & Temporal & Two-stream \\
\hline
Local branch only & 87.4\% & 87.3\% & 94.0\% \\
\hline
Non-local branch only (R-LSTM) & 86.9\% & 87.5\% & 93.8\% \\ 
\hline
Two-branch & \textbf{88.7\%} & \textbf{88.1\%} & \textbf{94.4\%} \\
\hline
\end{tabular}
\caption{Performance comparison of our architecture with only local branch vs. non-local branch vs. two-branch on UCF-101 (split 1).} 
\label{table:local-two}
\end{table}

\textbf{Effect of adding Relational LSTM block to different positions of ResNet-101-v2}: In this experiment, we aim to evaluate the impact of adding the Relational LSTM block to different positions of ResNet-101-v2. For the sake of still providing high-level features for Relational LSTM block and limiting computational cost, we mutate our architecture with Relation LSTM block added after conv5\_1 layer, conv5\_2 layer and conv5\_3 layer of ResNet-101-v2. The results are shown in Table \ref{table:adding-block}. From the table, we can see that there is performance increase when adding Relational LSTM block after conv5\_2 layer instead of conv5\_1 layer, indicating inputs of higher-level features to the Relational LSTM block would help it better learn long-term temporal dynamics and object interaction features. However, the performance decreases when adding Relational LSTM block after conv5\_3 layer (right before spatial average pooling layer) instead of conv5\_2 layer. One possible explanation is that the positional information preserved through Relational LSTM block is not further explored when the block is added right before a spatial average pooling layer.

\begin{table}[htb]
\centering
\begin{tabular}{|c|c|c|c|}
\hline
Positions for adding Relational LSTM block & Spatial & Temporal & Two-stream \\
\hline
After conv5\_1 layer & 87.5\% & 87.7\% & 94.0\% \\
\hline
After conv5\_2 layer & \textbf{88.7\%} & \textbf{88.1\%} & \textbf{94.4\%} \\ 
\hline
After conv5\_3 layer & 87.7\% & 87.5\% & 93.8\% \\
\hline
\end{tabular}
\caption{Performance comparison of our architecture with adding Relational LSTM block to different positions of ResNet-101-v2 backbone on UCF-101 (split 1).} 
\label{table:adding-block}
\end{table}

\subsection{Comparison with related work}
\begin{figure}[tb]
\begin{center}
\mbox{
    \includegraphics[width=\linewidth]{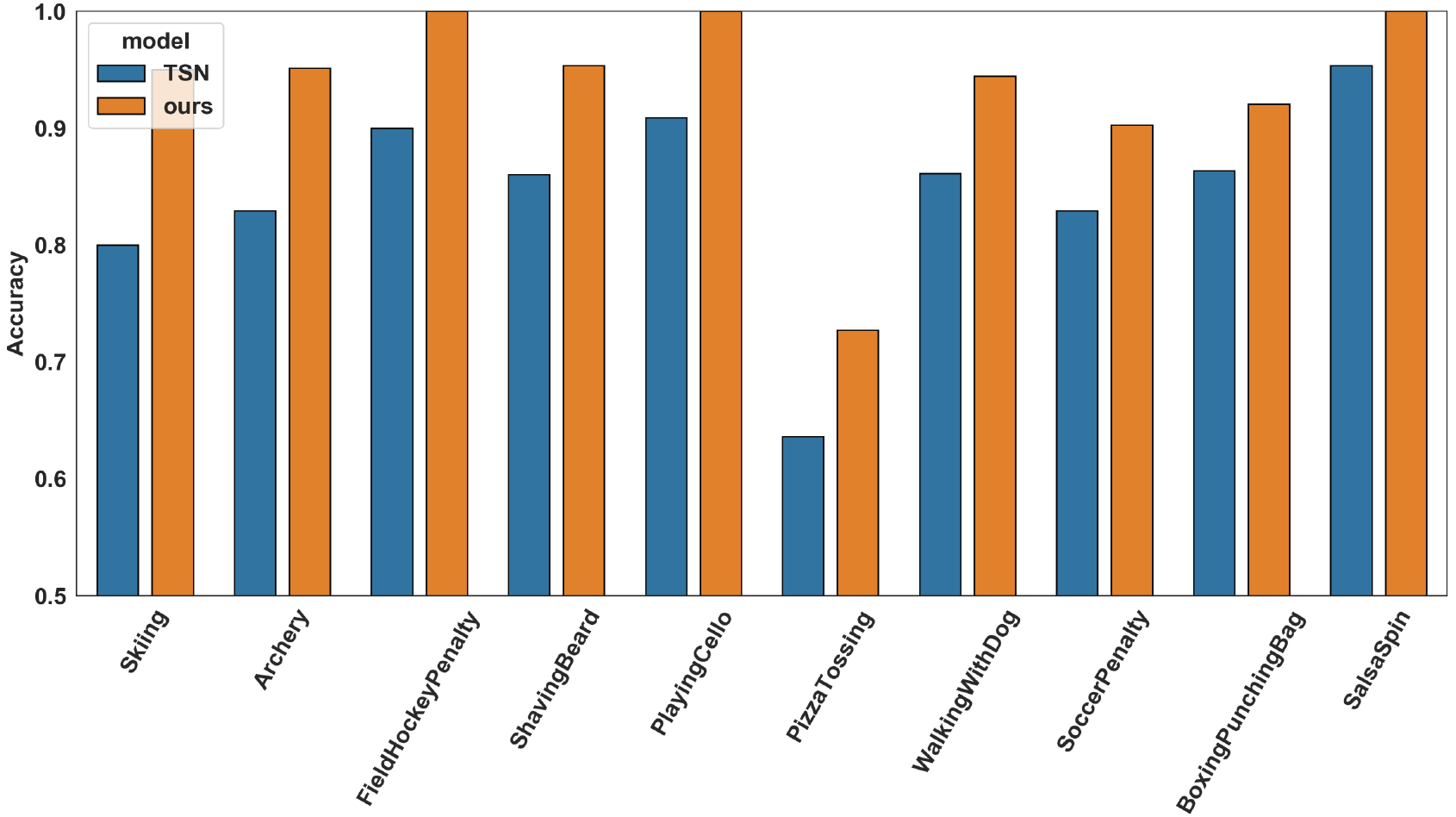}
  }
\end{center}
   \caption{10 classes of UCF-101 (split 1) with largest improvements from TSN to our two-branch architecture.}
\label{fig:visualization}
\end{figure}

\begin{figure*}[!ht]
\centering
\mbox{
    \includegraphics[keepaspectratio=true,scale=0.5]{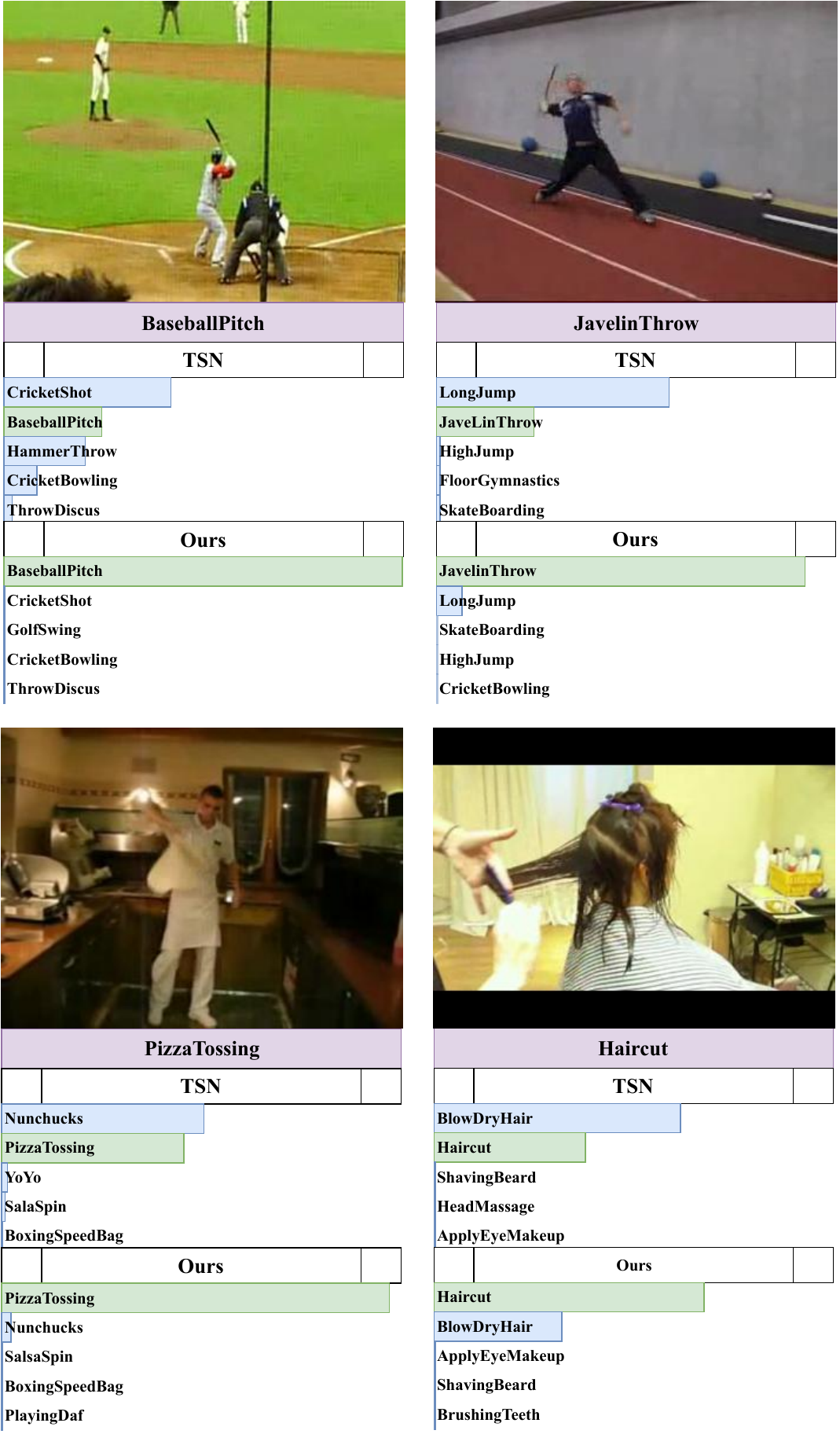}
  }
   \caption{\textbf{Visual comparison.} A visual comparison of top-5 predictions between TSN with our architecture on some instances of UCF-101 (split 1) test data. The text within the purple bar indicates the ground truth label of the instance, and that within the green/blue bars indicates correct/incorrect predictions. Lengths of the bars correspond to prediction probabilities.}
\label{fig:visual}
\end{figure*}

\textbf{Comparison with Temporal Segment Networks}: First, we compare our architecture with TSN on split 1 of UCF-101 dataset. The reason we compare with TSN is that our local branch is most similar to TSN, the only difference being that our local branch aggregates snippet-level features before FC layer, whereas TSN aggregates snippet-level prediction scores after FC layer. The performance of our implementation of TSN is shown in Table \ref{table:r-lstm vs tsn}, and the performance of our architecture is shown in Table \ref{table:local-two}. We observe that there is a 0.6\% increase in overall performance. In Figure~\ref{fig:visualization}, we show the 10 classes of UCF-101 with the largest improvements in our architecture over TSN. Most of those classes involve object interaction features, e.g. PizzaTossing, Archery, implying that the introduction of non-local branch reaps benefits towards relation reasoning. We also visualize some instances of UCF101 (split 1) test data in Figure~\ref{fig:visual} to show the importance of learning object interactions and long-term motion features. These examples involve object interactions and cannot be easily recognized using solely appearance features and short-motion features. For example, the instance about JavelinThrow has been misclassified as LongJump by TSN as both classes have the `running' sub-action in the early stages, whereas our architecture classifies its label correctly by a large margin. The possible reason is that our architecture learns about the interactions between the person and the javelin through time, but TSN erroneously focuses more on the running person. 

%%%%%%%%%%%%%%%%%%%%%%%%%%%
%%%%%%%%%%%%%%%%%%

%%%%%%%%%%%%%%%%%%%%%%%%%%%
%%%%%%%%%%%%%%%%%%

\setlength{\tabcolsep}{1pt}
\begin{table}[htb]
\begin{center}
\begin{tabular}{|c|C{2.5cm}|C{2.5cm}|}
\hline
Methods & UCF-101 & HMDB-51\\ 
\hline
Two-Stream+LSTM~\cite{ng2015beyond} & 88.6\% & - \\
VideoLSTM~\cite{li2018videolstm} & 89.2\% & 56.4\%\\
HAN~\cite{wang2016hierarchical} & 92.7\% & 64.3\%\\
L$^2$STM~\cite{sun2017lattice} & 93.6\% & 66.2\%\\
\hline
R-LSTM(ours) & 94.2\% & - \\
Two-branch(ours) & \textbf{94.8\%} & \textbf{71.4\%}\\
\hline
\end{tabular}
\end{center}
\caption{Comparison with LSTM-based state-of-the-art architectures on UCF-101 and HMDB-51 datasets. The performance accuracy is reported over all three splits.}
\label{table:lstm-sota}
\end{table}

\textbf{Comparison with LSTM-based state-of-the-art methods}: 
As our method belongs to the family of LSTM-based methods, we compare our method with other LSTM-based methods over all three splits of UCF-101 and HMDB-51, as shown in Table \ref{table:lstm-sota}. From the table, we observe that our method outperforms other LSTM-based methods by a large margin. Even with the standalone Relational LSTM network (non-local branch), over three splits of UCF-101, we achieve 94.2\% accuracy, which convincingly outperforms other LSTM-based methods. To the best of our knowledge, we achieve the best performance among all LSTM-based methods.

\setlength{\tabcolsep}{1pt}
\begin{table}[htb]
\begin{center}
\begin{tabular}{|c|C{2.5cm}|C{2.5cm}|}
\hline
Methods & UCF-101 & HMDB-51\\ 
\hline
iDT~\cite{wang2013action} & 86.4\% & 61.7\%\\
Two-Stream~\cite{simonyan2014two} & 88.0\% & 59.4\%\\
KVMDF~\cite{zhu2016key} & 93.1\% & 63.3\%\\
ST-ResNet~\cite{feichtenhofer2016spatiotemporal} & 93.4\% & 66.4\%\\
Two-Stream I3D~\cite{carreira2017quo} & 93.4\% & 66.4\%\\
TSN~\cite{wang2016temporal} & 94.0\% & 68.5\% \\
ST-Multiplier~\cite{feichtenhofer2017spatiotemporal} & 94.2\% & 68.9\%\\
ST-Pyramid Network~\cite{wang2017spatiotemporal} & 94.6\% & 68.9\%\\
MiCT-Net~\cite{zhou2018mict} & 94.7\% & 70.5\% \\
CoViAR~\cite{wu2017compressed} & 94.9\% & 70.2\%\\
\hline
Ours & \textbf{94.8\%} & \textbf{71.4\%}\\
\hline
\end{tabular}
\end{center}
\caption{Comparison with non-LSTM-based state-of-the-art methods on UCF-101 and HMDB-51 datasets. The performance accuracy is reported over all three splits. For a fair comparison, we only consider models that are pre-trained on ImageNet. We consistently set 0.45 for spatial and 0.55 for temporal stream in late fusion over the three splits of UCF-101 dataset and set 0.33 for spatial and 0.67 for temporal stream in late fusion over the three splits of HMDB-51 dataset.}
\label{table:sota}
\end{table}

\textbf{Comparison with non-LSTM-based state-of-the-art methods}
Before our work, LSTM-based methods have fallen behind non-LSTM-based methods in performance for a long period. So we provide comparisons of our method with current non-LSTM-based state-of-the-art methods over all three splits of UCF-101 and HMDB-51 datasets. We report these results in table \ref{table:sota}. From the table, we can observe that we obtain performance comparable to the top tier of existing state-of-the-art methods. Moreover, our method is simple and the resulting networks very easy to train. 

\setlength{\tabcolsep}{1pt}
\begin{table*}[htb]
\begin{center}
\begin{tabular}{|c|C{3.cm}|C{2.5cm}|}
\hline
Methods & Modality & Charades\\ 
\hline
Two-Stream~\cite{simonyan2014two} & RGB + flow & 18.6\%\\
Two-Stream + LSTM~\cite{sigurdsson2017asynchronous} & RGB + flow & 17.8\%\\
Asyn-TF~\cite{sigurdsson2017asynchronous} & RGB + flow & 22.4\%\\
CoViAR~\cite{wu2017compressed} & RGB & 21.9\%\\
MultiScale TRN~\cite{zhou2017temporal} & RGB & 25.2\%\\
TSN*~\cite{wang2016temporal} & RGB & 25.6\%\\
\hline
Two-branch(ours) & RGB & \textbf{28.8\%}\\
\hline
\end{tabular}
\end{center}
\caption{Comparison with the state-of-the-art methods on Charades dataset. The classification \textbf{mAP} is reported as evaluation metric. For a fair comparison, only methods using 2D ConvNets backbone are reported. ``*'' indicates our re-implementation of the method. }
\label{table:charades}
\end{table*}

\subsection{Experimental evaluation on large-scale Charades dataset}
To further evaluate the effectiveness of our method, we conduct experiments on it using the large-scale and complex Charades dataset~\cite{sigurdsson2016hollywood}. The Charades dataset demands more spatial and temporal relation reasoning by asking the actors to perform a sequence of actions involving interactions with objects in specific scenes. Following the same training and evaluation schema~\cite{sigurdsson2016hollywood}, we train our model on the actions extracted from 7,985 training videos and perform testing on the 1,863 validation videos. To perform inference on the entire video as a multi-label classification task, we sample 10 clips of 256 frames with equal temporal spacing from the video and evaluate each clip separately in the same way as mentioned in Section \ref{sec:implementation-details}. The predicted logits before softmax normalization are considered as the prediction scores for each clip and the video-level prediction scores are obtained by aggregating clip-level prediction scores with max pooling. Following most recent works~\cite{wu2017compressed, zhou2017temporal}, we perform experiments with only RGB images as inputs (spatial stream) for this dataset. For a fair and consistent comparison with state-of-the-art methods, we compare our model performance with methods that use 2D ConvNets as the backbone, as shown in Table \ref{table:charades}. As we can see from the table, our method outperforms other state-of-the-art methods by a large margin with only sparsely sampled RGB images as inputs. Furthermore, the performance difference between TSN and our method can validate the capability of our method for learning long-term temporal features and object interaction features.

\section{Conclusions and Future Work}
\label{sec:conclusion}
In this paper, we present a novel Relational LSTM module which we embed into a two-branch architecture for relation reasoning across space and time between objects in videos. It complements most existing action recognition methods for their lack of relation reasoning and learns video-level representations implicitly for modeling long-term trajectory features. In our experiments, we validate the contributions of introducing Relational LSTM module, and demonstrate the performance of our architecture on three challenging action recognition datasets. Before our work, LSTM-based methods lagged behind non-LSTM-based methods in performance, especially those that use I3D convolutions \cite{carreira2017quo}. Our method establishes state-of-the-art results among LSTM-based competitors and even enjoys performance comparable to non-LSTM-based counterparts on UCF-101 and HMDB-51 datasets. Moreover, our method achieves higher performance compared to the state-of-the-art methods that use 2D ConvNet backbones on the large-scale and complex Charades dataset. Even though we focus on the action recognition task in our experiments, much like the non-local block of Wang $et$ $al.$ \cite{wang2018non}, our Relational LSTM module can be inserted into various network architectures for other tasks. 
%In this work, we only explore the application of our Relational LSTM module in two-stream ConvNets. In the future, we will explore the possibility of applying it on 3D ConvNets. 

% add one sentence saying that our method can work on multi-label task

\section*{References}

\bibliography{mybibfile}

\end{document}